\def\year{2022}\relax
\documentclass[letterpaper]{article} 
\usepackage{aaai22}  
\usepackage{times}  
\usepackage{helvet}  
\usepackage{courier}  
\usepackage[hyphens]{url}  
\usepackage{graphicx} 
\usepackage{natbib}  
\usepackage{caption} 
\DeclareCaptionStyle{ruled}{labelfont=normalfont,labelsep=colon,strut=off} 
\usepackage{algorithm}
\usepackage{algorithmic}

\usepackage{newfloat}
\usepackage{listings}
\floatstyle{ruled}
\newfloat{listing}{tb}{lst}{}
\floatname{listing}{Listing}

\usepackage{todonotes}
\usepackage{newtxmath}
\usepackage{color}
\usepackage{subfigure} 
\graphicspath{{fig/}}

\usepackage{booktabs}
\usepackage{multirow}
\usepackage{longtable}
\newcommand{\captionskip}{0ex}
\newcommand{\figureskip}{0ex}
\newcommand{\tabelarraystretch}{1}

\title{Uncertainty-Aware Learning Against Label Noise on Imbalanced Datasets}
\author{Yingsong Huang\textsuperscript{\rm 1}\equalcontrib, Bing Bai\textsuperscript{\rm 1}\equalcontrib, Shengwei Zhao\textsuperscript{\rm 1}, Kun Bai\textsuperscript{\rm 1}, Fei Wang\textsuperscript{\rm 2}}
\affiliations{
\textsuperscript{\rm 1}Tencent Security Big Data Lab, Tencent Inc., China\\
\textsuperscript{\rm 2}Department of Population Health Sciences, Weill Cornell Medicine, USA\\
\{hudsonhuang,icebai,swzhao,kunbai\}@tencent.com\\
few2001@med.cornell.edu
}

\begin{document}

\maketitle

\begin{abstract}

Learning against label noise is a vital topic to guarantee a reliable performance for deep neural networks.
Recent research usually refers to dynamic noise modeling with model output probabilities and loss values, and then separates clean and noisy samples.
These methods have gained notable success. However, unlike cherry-picked data, existing approaches often cannot perform well when facing imbalanced datasets, a common scenario in the real world.
We thoroughly investigate this phenomenon and point out two major issues that hinder the performance, i.e., \emph{inter-class loss distribution discrepancy} and \emph{misleading predictions due to uncertainty}.
The first issue is that existing methods often perform class-agnostic noise modeling. 
However, loss distributions show a significant discrepancy among classes under class imbalance, and class-agnostic noise modeling can easily get confused with noisy samples and samples in minority classes.
The second issue refers to that models may output misleading predictions due to epistemic uncertainty and aleatoric uncertainty, thus existing methods that rely solely on the output probabilities may fail to distinguish confident samples. 
Inspired by our observations, we propose an Uncertainty-aware Label Correction framework~(ULC) to handle label noise on imbalanced datasets. 
First, we perform epistemic uncertainty-aware class-specific noise modeling to identify trustworthy clean samples and refine/discard highly confident true/corrupted labels.
Then, we introduce aleatoric uncertainty in the subsequent learning process to prevent noise accumulation in the label noise modeling process. We conduct experiments on several synthetic and real-world datasets. The results demonstrate the effectiveness of the proposed method, especially on imbalanced datasets.

\end{abstract}

\section{Introduction}

Widely accessible curated datasets, such as ImageNet~\cite{deng2009imagenet}, is one of the crucial factors that deep neural networks~(DNNs) flourished.
These datasets are usually artificially balanced among classes and carefully annotated by humans.
Nevertheless, high-quality human annotation is often expensive and time-consuming.
To obtain large-scale annotated data with limited budgets, practical solutions, such as crowd-sourcing~\cite{welinder2010multidimensional}, search engines~\cite{fergus2010learning}, or machine generation~\cite{kuznetsova2020open} are available.
However, these methods inevitably introduce mistakes or label noise, which in turn deteriorates the performance of DNNs~\cite{frenay2013classification}.
Another common problem in real-life applications is that some classes have a higher volume of examples in the training set than other classes, which we refer to as the class imbalance problem~\cite{buda2018systematic}. \citet{japkowicz2002class} had well studied that the class imbalance problem can significantly affect training machine-learning models.
When encountering class imbalance, existing label noise learning methods face new challenges, and how to deal with label noise on imbalanced datasets is essential in practice.

The effectiveness of recent works on noise-robust learning is often attributed to the \emph{memorization effect}~\cite{arpit2017closer,li2020gradient}, which states that DNNs learn patterns of correctly-labeled data first and then eventually memorize incorrectly-labeled data. 
On these grounds, based on the given noisy labels and the model's output probabilities, existing methods usually treat samples with small loss values in the early stage of model training as clean ones and then perform loss correction~\cite{reed2014training,konstantinov2019robust} or label correction~\cite{arazo2019unsupervised,tanaka2018joint}.

However, despite their immense success, these methods still face the following issues, especially in class imbalance situations.
Firstly, on imbalanced datasets, the loss distributions among classes show a significant discrepancy.
As pointed out by \citet{kang2019decoupling}, the norms of the classifier layer's weights correlate with the cardinality of the classes. Thus, the predicted probability of majority classes may naturally get more significant than that of minority classes, resulting in the inter-class loss distribution difference.
Existing methods that do not consider this difference may easily confuse samples in minority classes and samples with corrupted labels as their loss values all tend to be relatively large, thus getting bad performance for minority classes.

Secondly, models may be uncertain with their predictions, even if they give high output probabilities~\cite{gal2016uncertainty,gal2016dropout}.
Specifically, the wide range of possibilities in model structures and parameters leads to \emph{epistemic uncertainty}, while the uncertainty caused by noisy data is referred to as \emph{aleatoric uncertainty}~\cite{kendall2017uncertainties}. For learning against label noise, on the one hand, misleading predictions due to \emph{epistemic uncertainty} may affect the identification of clean samples from noisy ones, especially for the minority classes because models see much fewer samples in the minority classes and may get underfit during training. 
On the other hand, the label noise can hardly be eliminated. The residual noise may accumulate and introduce the \emph{aleatoric uncertainty} to the model, especially for the majority classes and during the late learning phase, where overfitting is more likely to occur.
These two kinds of uncertainty may impair the effectiveness of learning against label noise.

To address the shortcomings of existing methods on imbalanced datasets, we introduce a novel \underline{U}ncertainty-aware \underline{L}abel \underline{C}orrection~(ULC) framework. 
To better distinguish clean samples and correct labels in imbalanced noisy datasets, we improve the common class-agnostic practice by considering the inter-class loss distribution discrepancy problem and modeling samples' noise individually for each class. 
Furthermore, we estimate the model's epistemic uncertainty and incorporate the epistemic uncertainty in addition to the output probability, which establishes the proposed \underline{E}pistemic \underline{U}ncertainty-aware \underline{C}lass-\underline{S}pecific noise modeling method~(EUCS).
On the other side, as samples filtering and label correction can hardly be perfect, there is always residual noise that may bring negative impacts.
We propose the \underline{A}leatoric \underline{U}ncertainty-aware \underline{L}earning~(AUL) against label noise by estimating the sample-wise aleatoric uncertainty induced by the residual noise and adding corresponding logit corruption in the loss to prevent overfitting.

The key contributions of this work are summarized as follows.
\begin{itemize}
\item We analyze why existing methods for label correction may not work well with imbalanced datasets and ascribe the reasons for the discrepancy of loss distribution among classes and the ignorance of uncertainty.
\item We propose a novel uncertainty-aware label correction framework for imbalanced datasets.
The framework is based on a combination of epistemic uncertainty-aware class-specific noise modeling and aleatoric uncertainty-aware learning against label noise.
\item Extensive experiments on synthetic and real-world noisy datasets show the effectiveness of the proposed framework, especially on imbalanced datasets.
\end{itemize}

\section{Related work}

In this section, we review important related work, including learning with noisy labels, and the methods for uncertainty estimation.

\paragraph{Learning with noisy labels}
Recent studies on noise-robust learning with DNNs primarily fall into two categories, namely \emph{loss correction} and \emph{label correction}.

Loss correction methods either explicitly rectify the cost function considering the noise distribution~\cite{sukhbaatar2014learning,patrini2017making,hendrycks2018using,han2018masking}, or propose loss functions specifically designed to weaken the impact of noisy samples~\cite{ghosh2017robust,zhang2018generalized,ma2018dimensionality}.
On the other side, recent research focuses more on label correction which deals with label noise by identifying and correcting noisy labels. 
For examples, \citet{tanaka2018joint} correct labels as model predictions~(soft labels) or one-hot vectors~(hard labels) during training with an alternating optimization approach. 
\citet{arazo2019unsupervised} model label noise with a two-component mixture model on the loss to discriminate noise from clean samples, correct labels as a convex combination of noisy labels and predicted soft or hard labels, and apply MixUp~\cite{zhang2017mixup} to enhance performance. 
DivideMix~\cite{li2020dividemix} applies MixMatch~\cite{berthelot2019mixmatch} in a semi-supervised manner and exploits Co-teaching~\cite{han2018co} to avoid confirmation bias. 
\citet{nishi2021augmentation} explores different augmentations for noise modeling and learning in the label correction framework, improving results on the state of the art.
SMP~\cite{han2019deep} generates corrected labels by feature clustering and multi-prototypes selection, which is an elaborate process.


Our proposed approach falls into the \emph{label correction} category.
Although previous label correction approaches have achieved success on nearly class-balanced noisy datasets, the performance on class-imbalanced datasets has not been deeply studied.
Apart from methods that focus on learning against label noise,
\citet{koziarski2020combined} and \citet{chen2021rsmote} propose to improve the Synthetic Minority Over-sampling Technique~(SMOTE)~\cite{chawla2002smote} under label noise. However, their primary goal is to perform better over-sampling against the class imbalance problem, so their focus is different from ours.

\paragraph{Uncertainty estimation}
Uncertainty estimation to establish prediction confidence is vital for deep learning and has been studied thoroughly~\cite{lakshminarayanan2016simple,gal2016uncertainty,malinin2018predictive,maddox2019simple}.
Bayesian Neural Networks~(BNNs)~\cite{gal2016dropout} capture epistemic uncertainty in neural networks by placing a prior distribution over the model's parameters. 
\citet{kendall2017uncertainties} present a framework combining input-dependent aleatoric uncertainty together with epistemic uncertainty.
In addition, for applications, uncertainty has been applied to improve the performance of semi-supervised learning~(SSL) and self-training, and achieves desired effects~\cite{rizve2020defense,mukherjee2020uncertainty}.

However, learning against noisy labels on imbalanced datasets is more challenging, as it requires disambiguation of epistemic uncertainty from aleatoric uncertainty~\cite{northcutt2021confident}. 
In this paper, we address this problem and propose the ULC framework.

\section{Perspective}
As the proposed ULC falls into the label correction category, this section presents our perspective regarding the issues of existing label correction frameworks when dealing with class-imbalanced datasets.
We briefly introduce the label correction framework in deep learning, discuss the issues, and present case studies.

\subsection{Label Correction Frameworks Against Label Noise}
Let $\mathcal D = ( \mathcal X,\widetilde{\mathcal Y}) = \{(x_i,\widetilde{y}_i)\}_{i=1}^N$ be a noisy dataset with $N$ samples, where $x_i$ is the input and $\widetilde{y}_i$ is the associated one-hot label over $C$ classes with possible noise. We denote the latent true label corresponding to $\widetilde{y}_i$ by $y^*_i$ such that $\widetilde{y}_i,y^*_i \in \{0,1\}^C$ and $\mathbf{1}^\top \widetilde{y}_i = 1, \mathbf{1}^\top y^*_i = 1$. In a more general case, there is $p(\widetilde{y}_{ij} = 1|{y^*_{ik}} = 1,x_i) = {\eta}^{x_i}_{jk}$, where $\widetilde{y}_{ij}$ corresponds to the $j$-th element of $\widetilde{y}_i$ and $y^*_{ik}$ corresponds to $k$-th element of $y^*_i$. 
Let $f_{W}(x) = v$ denote the logits vector for $x$, where $f_{W}$ is a DNN model with trainable parameters $W$. 
The final class prediction $\widehat{y}$ is squashed by a softmax function, such that $\widehat{y} = \textrm{softmax}(v)$. 
For cross-entropy loss $\ell$, which is the most commonly used loss for classification, the empirical risk is as follows:
\begin{equation}
\label{eq:emprical_risk}
\small
    \ell(W) = \frac{1}{N}\sum_{i=1}^{N}\ell_i = -\frac{1}{N}\sum_{i=1}^{N}\widetilde{y}_i^\top \log(\widehat{y}_i)\,,
\end{equation}
where $\ell_i = \ell_i(W)$, and $\widehat{y}_i = \widehat{y}_i(x_i;W)$.

On account of noisy labels $\widetilde{y}_i$, the empirical risk minimization under loss $\ell$ leads to an ill-suited solution. Label correction methods refine/discard the observed true/corrupted labels $\widetilde{y}_i$ and learn on the corrected labels. Generally, we denote the clean samples by $x_i \in \mathcal{\widehat{X}}$ with corrected labels denoted by $y_i \in \mathcal Y$, and denote the rejected noisy samples that we cannot assign any labels confidently by $x_i \in \mathcal{\widehat{U}}$. Accordingly, the optimization problem concerning learning on noisy labels with label correction is formulated as semi-supervised learning~(SSL), which is as follows:
\begin{equation}
\begin{aligned}
\label{eq:ssl}
\small
  \min_{W,\mathcal{Y}} \ell_c &= \ell_{x}(W) + \lambda_u \ell_{u}(W) + \lambda_r \ell_{\textrm{reg}}(W)\\
  \ell_{x}&(W) = -\frac{1}{\left| \mathcal{\widehat X}\right|}\sum_{x_i \in \mathcal{\widehat{X}}}y_i^\top \log(\widehat{y}_i)\,,
\end{aligned}
\end{equation}
where $y_i$ can be initialized with $\widetilde{y}_i$, and update during model training, $\ell_u$ denotes any unsupervised learning loss, and $\ell_\textrm{reg}$ denotes any regularization term.

Alternating optimization methods have turned to be adequate to progressively search for a proper solution for $\mathcal Y$ and $W$, in which $\mathcal Y$ can be estimated based on the predicted probability $\widehat{y}_i$, the loss value $\ell_i$ or the feature representations~\cite{tanaka2018joint,li2020dividemix,han2019deep}. It appears to be more robust to establish $\mathcal Y$ in terms of the feature representations. However, it is too complicated to select valid prototypes.
Alternatively, existing methods often update $\mathcal Y$ based on loss values.
\begin{equation}
\small
    y^{n+1}_i = h(\mathcal{L}^{n},\widetilde y_i)\,,
\end{equation}
where $\mathcal{L}^{n} = \left \{{\ell_i}^{n}\right \}$ is the set of loss values in the $n$-th update and $\mathcal Y^{n+1} = \left \{y_i^{n+1} \right \}$ is corrected labels in the ($n$+1)-th update.

For instance, recent works~\cite{arazo2019unsupervised,li2020dividemix} conceive of the distribution of loss values $\mathcal{L}$ as a two-component Beta Mixture Model~(BMM) or Gaussian Mixture Model~(GMM), and then refine the labels with:
\begin{equation}
\small
    y^{n+1}_i = h(\mathcal{L},\widetilde y_i) = p(y^*_i = \widetilde{y}_i|\mathcal{L})\widetilde{y}_i + (1-p(y^*_i = \widetilde{y}_i)|\mathcal{L}))\widehat{y}_i\,,
\end{equation}
where $p(y^*_i = \widetilde{y}_i|\mathcal{L})$ is the estimated probability of the observed label being true, corresponding to the posterior probability of the BMM or GMM component with the smaller mean~(smaller loss).

\subsection{Rethinking Label Correction Frameworks}
In this section, we analyze why existing methods may fail to perform well under class imbalance.
Existing noise modeling methods ignore that 1)~the loss distribution for each class may vary significantly, especially on class-imbalanced data; 2)~due to the existence of uncertainty, the predictive probability may fail to indicate the confidence accurately.

\paragraph{Inter-class loss distribution discrepancy}
\citet{kang2019decoupling} decouple representations and classifiers in the learning procedure to reveal the effect of class imbalance.
The work shows that the norms of the classifier layer's weights are correlated with the cardinality of the classes. 
When trained on imbalanced data, the inter-class loss distribution inevitably disagrees with each other.
Samples in majority classes get larger logits and smaller losses, while samples in minority classes get smaller logits and larger losses. 

\begin{figure}[t]
    \centering
    \includegraphics[width=0.75\columnwidth]{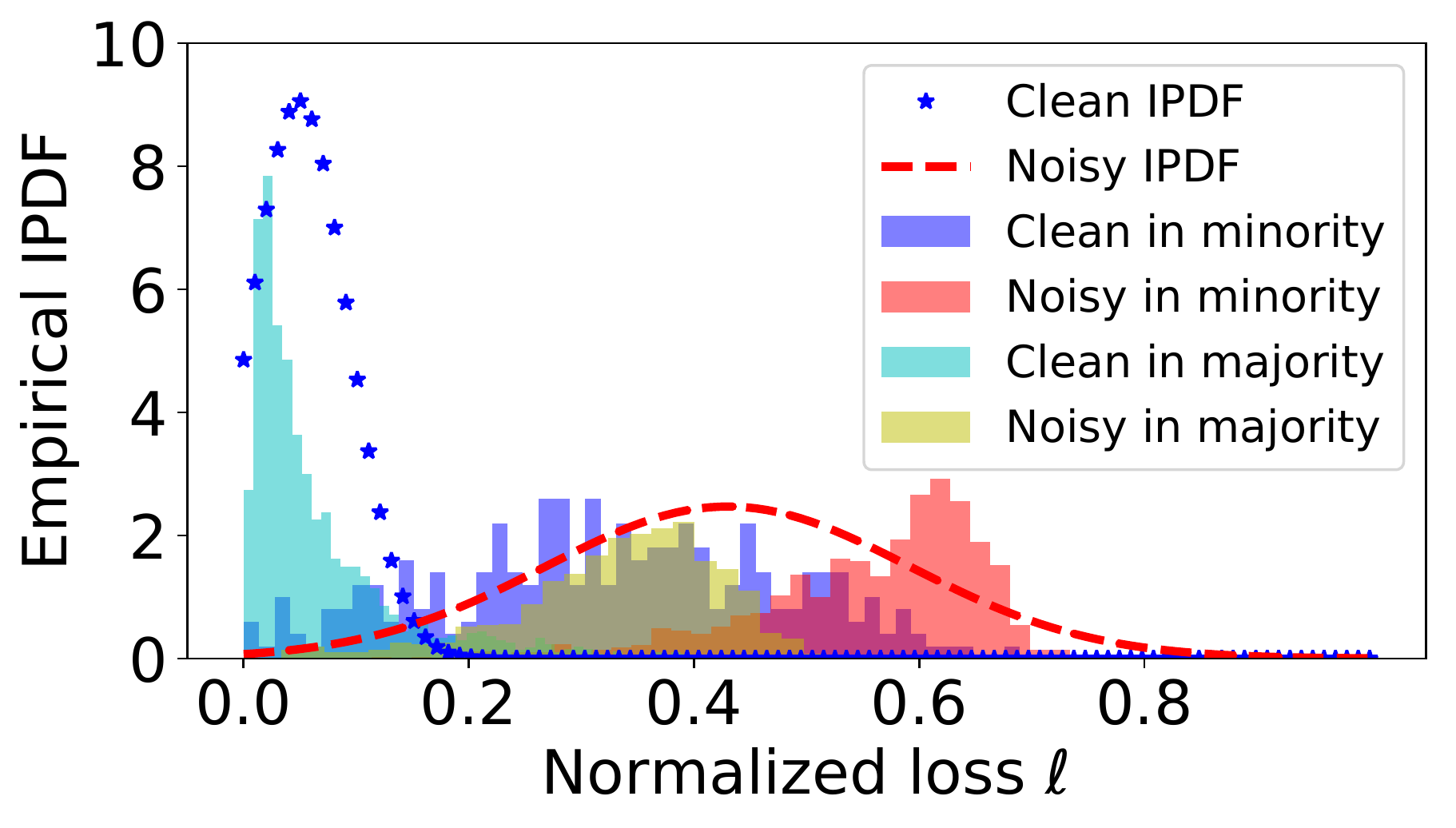}
    \vspace{\captionskip}
    \caption{A binary example for empirical IPDF and estimated with class-agnostic two-component GMM over loss values on noisy CIFAR-10, warm-up of 30 epochs. Under 50\% symmetric noise and 1:10 class imbalance, most of the samples in minority classes would be handled as noise. 
    Class-agnostic label noise modeling fails due to the discrepancy between inter-class loss distributions.}
    \label{Fig1}
    \vspace{\figureskip}
\end{figure}

Figure~\ref{Fig1} illustrates the problem mentioned above by presenting a binary example, which shows the individual probability density function~(IPDF) of clean and noise components modeled by GMM.
The case is under 50\% symmetric noise, we randomly choose two classes for demonstration, i.e., class~\#0 for the minority and class~\#3 for the majority, and they are under 1:10 imbalance.
As we can see, the loss distributions of noisy samples in the majority class overlaps with clean samples in the minority class.
With a two-component GMM, most of the samples with the observed label in the minority class are handled as noise.
These results suggest that class-agnostic noise modeling may not work well when the loss distributions of the different classes vary significantly, which can become a typical case with imbalanced datasets.

\paragraph{Misleading predictions due to uncertainty}
Misleading predictions exist, accompanied by a high level of epistemic uncertainty~\cite{gal2016uncertainty}.
Learning with noisy labels needs to overcome epistemic and aleatoric uncertainty simultaneously. Moreover, epistemic uncertainty and aleatoric uncertainty are dominant in different classes and phases during the learning procedure.

On the one hand, epistemic uncertainty due to the possibilities of model structures and parameters could be reduced by observing more data~\cite{gal2016dropout}. However, models may get underfit with minority classes, especially in the early learning phase which is critical for the effectiveness of the memorization effect. 
Thus epistemic uncertainty may have a more significant influence on minority classes.
On the other hand, aleatoric uncertainty induced by residual label noise may lead to overfitting to noisy samples~\cite{gal2016uncertainty}, especially for the majority classes and during the late learning phase.
Moreover, aleatoric uncertainty cannot be explained away simply with observing more data~\cite{kendall2017uncertainties}.

\begin{figure}[t]
    \centering
    \subfigure[Distribution of loss values]{
        \label{Fig2.a}
        \includegraphics[width=0.465\columnwidth]{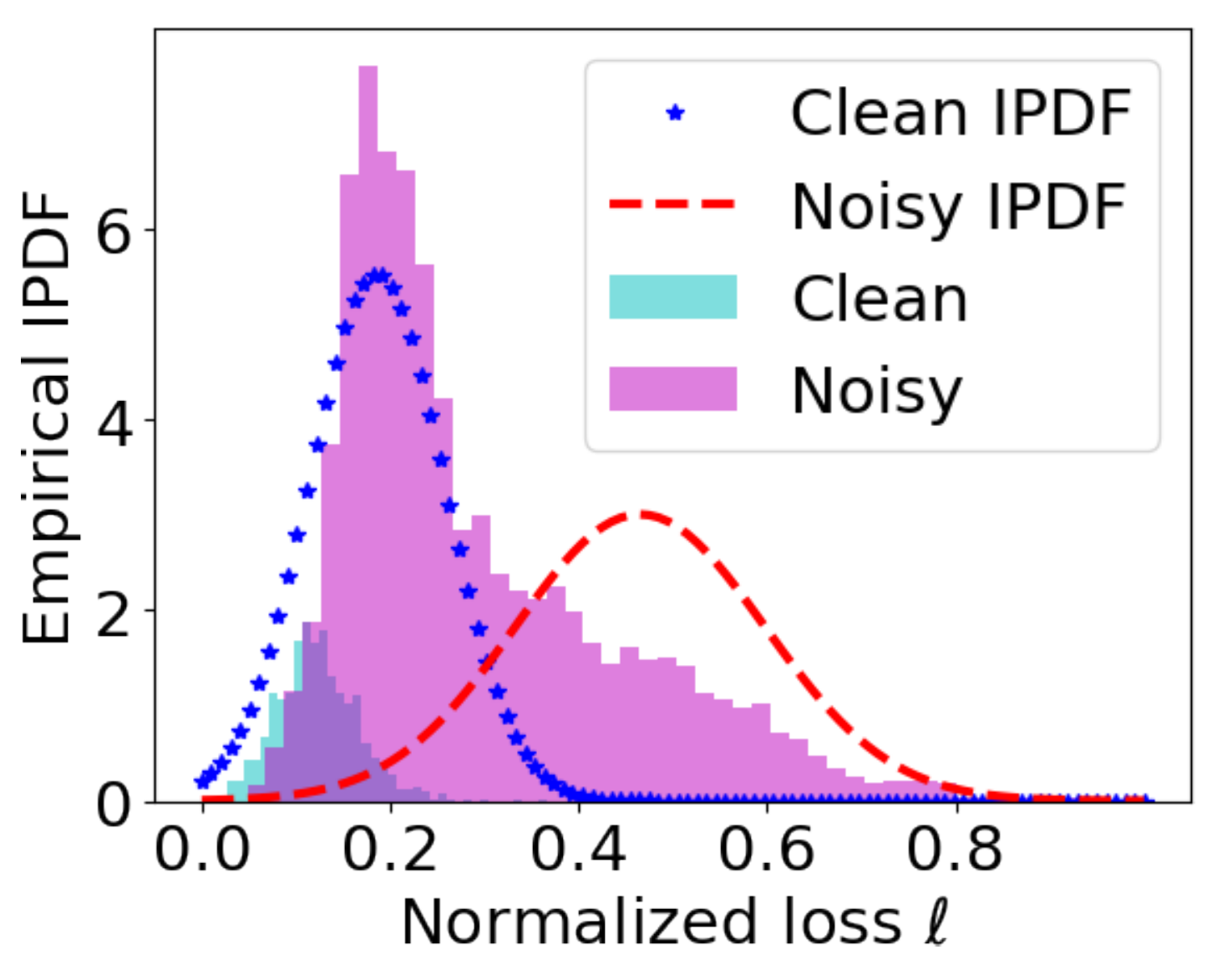}
        
    }
    \subfigure[Joint distribution of loss values and epistemic uncertainty]{
        \label{Fig2.b}
        \includegraphics[width=0.465\columnwidth]{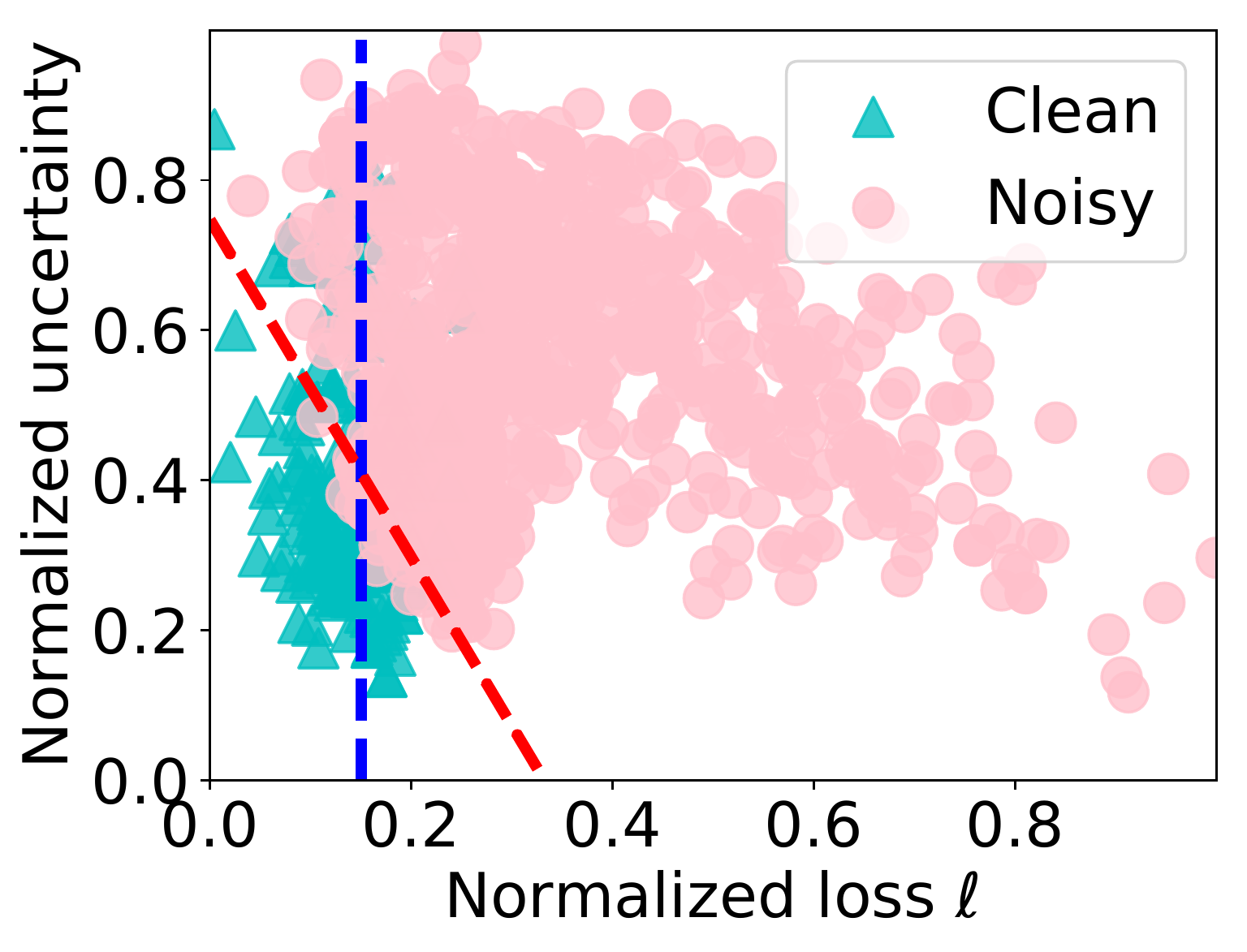}
    }
    \vspace{\captionskip}
    \caption{Loss values and epistemic uncertainty distribution on CIFAR-10 with 90\% symmetric noise, a warm-up of 30 epochs. (a)~Loss distribution of clean samples overlaps with noisy samples due to misleading predictions accompanied by a high level of epistemic uncertainty. (b)~Label noise modeling considering epistemic uncertainty distinguishes clean samples from noisy ones more accurately.}
    \label{Fig2}
    \vspace{\figureskip}
\end{figure}

In Figure~\ref{Fig2}, we present the loss values and epistemic uncertainty distribution on CIFAR-10 with 90\% symmetric noise after warming up 30 epochs. Figure~\ref{Fig2.a} shows that if we only use the loss to distinguish clean samples, the separability will be poor, as the loss distribution of clean samples overlaps with noisy samples due to misleading predictions accompanied by a high level of uncertainty. While if we turn to the joint distribution of loss values and epistemic uncertainty, we can find that label noise modeling considering epistemic uncertainty distinguishes clean samples from noisy ones much more accurately, as shown in Figure~\ref{Fig2.b}.

\section{Method}
In this section, to address the issues discussed above, 
we propose the epistemic uncertainty-aware class-specific noise modeling method to fit the inter-class discrepancy on loss distribution. 
Then we formulate aleatoric uncertainty as class-dependent and instance-dependent Gaussian noise for logit corruption to prevent noise accumulation.

\subsection{Epistemic Uncertainty-aware Class-specific Noise Modeling Adapted for Class-imbalanced Data}


Inspired by these observations of inter-class loss distribution discrepancy and misleading predictions accompanied by a high level of epistemic uncertainty, we propose a robust epistemic uncertainty-aware class-specific label noise modeling method.
We obtain the predicted values and epistemic uncertainty, then we integrate the results to assess whether the labels are true or corrupted.

We first estimate the integrated prediction and the epistemic uncertainty.
The basic idea is that instead of taking the model's parameters as fixed values, we place a prior distribution over the parameters $p(W)$, infer the posterior $p(W|\mathcal{X, \widetilde{Y}})$, and finally obtain the marginal probability $p(\widetilde{\mathcal{Y}}|\mathcal{X})$.
We then can estimate the epistemic uncertainty as the entropy of $p(\widetilde{{\mathcal{Y}}}|\mathcal{X})$.
This procedure can be processed efficiently by MC-Dropout~\cite{gal2016dropout}.
Considering $q_\theta(W)$ to be the surrogate Dropout distribution, we perform $T$ stochastic forward passes through the network with dropout enabled for each instance $x_i$.
Thus, with sampled model weights $\{W_t\}_{t=1}^T \sim q_\theta(W)$, the approximate integrated prediction can be obtained as $\widehat{y}_i = \frac{1}{T}\sum_{t=1}^{T}\textrm{softmax}(f(x_i,W_t))$
and the epistemic uncertainty $\epsilon_i=\epsilon(\widehat{y}_i)$ can be estimated using the entropy of predicted probability vector and then normalized\footnote{Readers may refer to the appendix and \citet{gal2016dropout} for more information about uncertainty estimation.}.

Then, we try to estimate the probability of the observed label being true, i.e., the clean probability $\omega_i=p(y^*_i=\widetilde{y}_i|\ell_i,\epsilon_i)$.
The posterior probability conditional on class-specific loss values and epistemic uncertainty cannot be evaluated analytically because the inference of joint distribution of loss values and epistemic uncertainty is intractable. 
However, we find that the posterior probability can be approximately negatively correlated to epistemic uncertainty~(see Figure~\ref{Fig2.b}).
We assume that the effects of epistemic uncertainty and loss values are independent of each other, and samples with lower-level epistemic uncertainty and smaller loss values are more presumably to be clean. 
Thus, we empirically approximate the posterior probability as the weighted geometric mean of $1-\epsilon_i$ and $p(y^*_i=\widetilde{y}_i|\ell_i)$: 
\begin{equation}
\label{eq:clean_proba}
\small
    \omega_i=p(y^*_i=\widetilde{y}_i|\ell_i,\epsilon_i)= (1-\epsilon_i)^{r} p(y^*_i=\widetilde{y}_i|\ell_i)^{1-r}\,,
\end{equation}
where $r$ is the hyper-parameter to balance the influence of epistemic uncertainty and loss values.

With the uncertainty $\epsilon_i$ estimated, the problem left is to obtain $p(y^*_i=\widetilde{y}_i|\ell_i)$.
Similar to \citet{li2020dividemix}, this probability can be approximated by fitting a two-component mixture model on loss values. 
However, considering the inter-class loss distribution discrepancy~(see Figure~\ref{Fig1}), we model the loss distribution separately for each class $j$. 
Let $\mathcal{L}_j = \{\ell_i|\ell_i \in \mathcal{L}, \widetilde{y}_{ij}=1\}$ be the set of loss values of instance with observed label $j$. 
We fit a GMM on $\mathcal{L}_j$, and the mean of the two components are denoted by $\mu_{j0}$, $\mu_{j1}$\footnote{We find that BMM easily degenerates under a high-level epistemic uncertainty.}. 
Generally, we assume $\mu_{j0} \le \mu_{j1}$.
Hence, the probalility $p(y^*_i=\widetilde{y}_i|\ell_i)$ can be approximated as $p(\mu_{j0}|\ell_i)$.

Finally, we identify the clean and noisy samples by placing a threshold $\tau$ on the probabilities $\omega_i$. 
We discard corrupted labels and mark the samples as unlabeled. True labels are also refined as:
\begin{equation}
\label{eq:rectified_label}
\small
    y_i = \omega_i \widetilde{y}_i + (1-\omega_i)\widehat{y}_i\,.
\end{equation}

\begin{figure}[t]
    \centering
    \subfigure[Comparison of different label noise modeling methods]{
        \label{Fig3.a}
        \includegraphics[width=0.465\columnwidth]{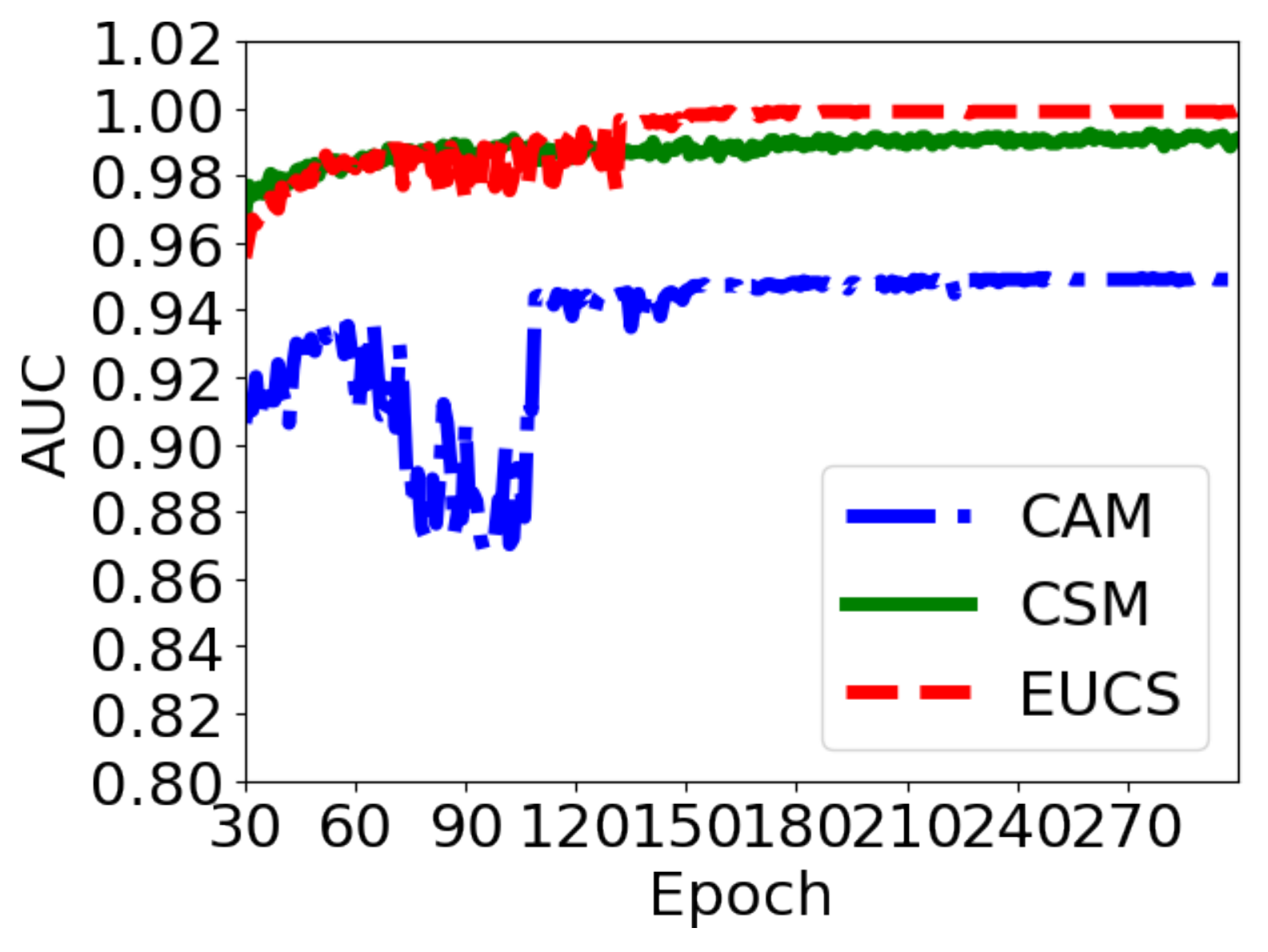}
    }
    \subfigure[Effects of different label noise modeling methods on minority and majority classes]{
        \label{Fig3.b}
        \includegraphics[width=0.465\columnwidth]{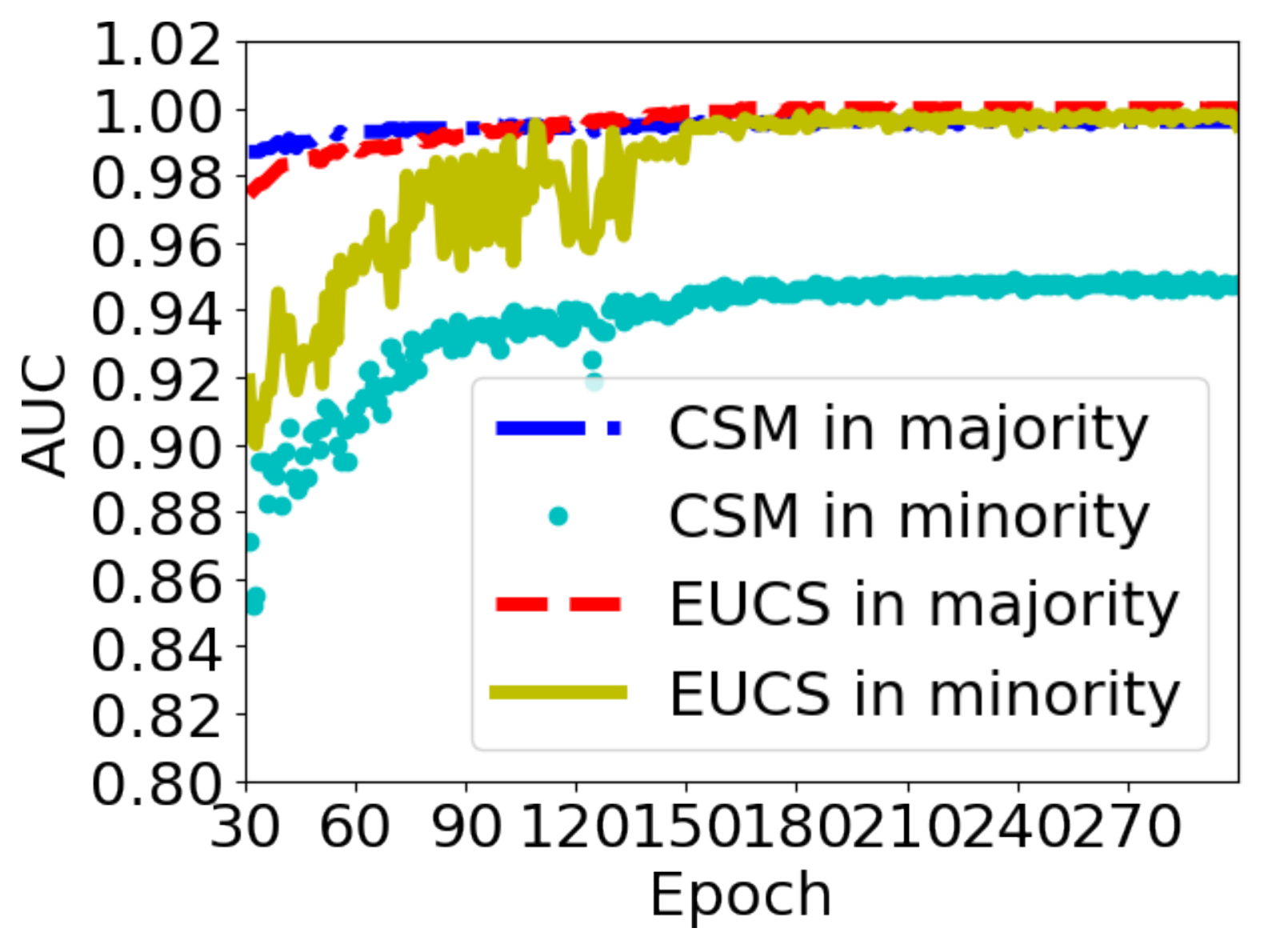}
    }
    \vspace{\captionskip}
    \caption{AUC for clean/noisy sample classification on synthetic CIFAR-10 under 1:10 imbalance with 50\% symmetric noise. Train for 300 epochs with different label noise modeling strategies. (a)~In contrast to Class-agnostic modeling~(CAM), Class-specific modeling~(CSM) is more robust to class imbalance and separates clean and noisy samples more accurately under class imablance. (b)~EUCS mainly contributes to identifying noise in minority classes compared to CSM.}
    \label{Fig3}
    \vspace{\figureskip}
\end{figure}

Figure~\ref{Fig3} shows the effectiveness of the proposed epistemic uncertainty-aware class-specific label noise modeling~(EUCS) method.
We evaluate the performance on synthetic CIFAR-10 and present the area under the receiver operating characteristic curve~(AUC) metrics for clean/noisy sample classification based on Eq.~(\ref{eq:clean_proba}).
We can find that EUCS consistently outperforms traditional class-agnostic modeling~(CAM) and class-specific modeling~(CSM) without considering epistemic uncertainty. 
Moreover, label noise modeling with consideration
of epistemic uncertainty can significantly facilitate the identification of noise in minority classes.

\subsection{Aleatoric Uncertainty-aware Learning Against Label Noise}
Noise modeling can filter out most of the label noise. However, residual noise still exists and contributes to overfitting.
To relieve misleading predictions induced by residual noise, we model latent aleatoric uncertainty in the learning procedure and introduce it as logit corruption to prevent overfitting.

Firstly, we model aleatoric uncertainty. 
\citet{kendall2017uncertainties} shows that aleatoric uncertainty can be modeled by logit corruption with Gaussian noise, and the formulation leads to new loss functions, which can be interpreted as learned loss attenuation – attenuating the effect from corrupted labels and making the loss more robust to noisy data.  
Inspired by this, we inject instance-dependent and class-dependent noise $\delta^{x_i}_{\mathcal Y}$ into the logit vector. 
We assume that the class-dependent and instance-dependent noisy factors in $\delta^{x_i}_{\mathcal Y}$ are independent of each other. 
Let $\delta^{x_i}$ be a diagonal matrix, we refer to it as the instance-dependent noise factor.
Let $\delta^ {\mathcal Y}$ be a stochastic matrix, such that $\sum_i{\delta^{\mathcal Y}_{ij}}=1$, we refer it as the class-dependent noise factor.
The induced logit corruption process can be formulated as:
\begin{equation}
\small
    \widehat{v}_i(W) = \delta^{x_i}(W) + {\delta^{\mathcal Y} } f_W(x_i)\ .
\end{equation}
We approximate $\delta^{\mathcal Y}$ by $I + \delta$.
In addition, we assume a Gaussian distribution over $\delta^{x_i}$ and $\delta$ with variance $\sigma^{x_i}$ and $\sigma$, i.e., $\delta^{x_i} \sim \mathcal N(0,\sigma^{x_i})$, $\delta \sim \mathcal N(0,\sigma)$. 
$\sigma^{x_i}$ and $\sigma$ can be viewed as a measure of aleatoric uncertainty.
The corrupted logits vector is squashed with the softmax function to obtain $\widehat y_i$:
\begin{equation}
\small
    \widehat y_i = \textrm{softmax}((I + \delta)f_W(W) + \delta^{x_i}(W))
\end{equation}

Then we estimate $\sigma^{x_i}$ and $\sigma$, and learn with loss attenuation.
We attach an affiliated branch to the head of the network to predict $\sigma^{x_i}$. 
The reparametrization trick~\cite{kingma2015variational} is used to obtain a minibatch-based Monte Carlo estimator of $\ell_x$. The numerically stable stochastic loss for cleaned samples is given by:
\begin{equation}
\label{eq:aul}
\small
    \ell_{x}(W) = -\frac{1}{\left | \mathcal{\widehat X} \right |}\sum_{x_i \in \mathcal{\widehat{X}}}(y_i)^\top \log\frac{1}{T}\sum_{t=1}^{T}(\widehat{y}_{it}(x_i;W,\sigma^{x_i},\sigma))
\end{equation}
with $\widehat y^{(t)}_i$ the softmax ouput of the $t$-th sampling over logits conditional on $\sigma^{x_i}$ and $\sigma$. The extra computation cost is marginal since that $\sigma$ is a $C \times C$ matrix and $\sigma^{x_i}$ is actually a $C \times 1$ vector and sampling on them is extremely fast.

\subsection{Implementation Details}
The algorithm procedure follows label correction frameworks similar to DivideMix~\cite{li2020dividemix}, and we add the proposed epistemic uncertainty-aware class-specific noise modeling and aleatoric uncertainty-aware learning.

We conventionally warm up the model for a few epochs using cross-entropy loss with an entropy term, then the learning procedure alternates between two phases.
First, as introduced before, we estimate the clean probabilities $\omega_i$ and the corrected labels $y_i$ by Eq.~(\ref{eq:clean_proba}) and (\ref{eq:rectified_label}).
With a threshold $\tau$ on $\{\omega_i\}$, we divide the training data into clean samples with corrected labels $\mathcal{\widehat X}$ and noisy samples without labels $\mathcal{\widehat U}$.
Following DivideMix, Co-teaching is used to alleviate the sample selection bias.
Next, semi-supervised learning~(SSL) is performed based on $\mathcal{\widehat X}$ and $\mathcal{\widehat U}$.
Also following DivideMix, we use MixMatch~\cite{zhang2017mixup} to augment $\mathcal{\widehat X} \cup \mathcal{\widehat U}$ into $\mathcal{X'} \cup \mathcal{U'}$.
During the SSL phase, as discussed before, we sample on logits conditional on aleatoric uncertainty $\delta^{x}$ and $\delta$, and introduce the aleatoric uncertainty as logit corruption to prevent overfitting by Eq.~(\ref{eq:aul}). 
Accordingly, the total stochastic loss in Eq.~(\ref{eq:ssl}) can be rewritten as:
\begin{equation}
\small
\begin{aligned}
    & \ell_c = \ell_x + \lambda_u \ell_u\\
    \ell_{x} =  -\frac{1}{\left | \mathcal{X'} \right |} \sum_{x_i,y_i \in \mathcal{X'}}&(y_i)^\top \log \frac{1}{T} \sum_{t=1}^{T}(\widehat{y}_{it}(x_i;W,\sigma^{x_i},\sigma)) \\
    \ell_u = \frac{1}{\left |\mathcal{U'} \right |} \sum_{x_i,y_i \in \mathcal{U'}}&{\left \|y_i - \frac{1}{T} \sum_{t=1}^{T}(\widehat{y}_{it}(x_i;W,\sigma^{x_i},\sigma))\right \|}^2_2\,.\\
\end{aligned}
\end{equation}

The additional regularization term $\ell_\textrm{reg}$ in Eq.~(\ref{eq:ssl}) used in prior label correction works~\cite{tanaka2018joint,arazo2019unsupervised,li2020dividemix} imposing a uniform prior distribution assumption on the prior probabilities is proven to be imperative to prevent the assignment of all labels to a single class. However, we find it inessential in our method.

In the appendix, we outlines the uncertainty-aware label correction process, and summarize the notations.

\section{Experiments}
In this section, we report the experimental results. We introduce the experimental setup, report the performance of ULC and baselines, and also analyze the results of ablation experiments.

\subsection{Experimental Setup}
\paragraph{Datasets}


We perform extensive evaluations on five datasets: CIFAR-10, CIFAR-100, class-imbalanced CIFAR-10, class-imbalanced CIFAR-100, and Clothing1M. 
The noise injection methods for CIFAR-10 and CIFAR-100 follow the previous work~\cite{li2020dividemix}. 
We experiment with two types of label noise: symmetric and asymmetric.
To test the robustness of our approach to class imbalance, we resample the training set to construct class-imbalanced CIFAR-10 and CIFAR-100.
Specifically, we randomly choose half the classes and randomly sub-sample 1/5 and 1/10 examples in these classes while other classes remain the same. 
Note that we only resample the training sets for class imbalance settings, i.e., the test sets remain the same as class balance settings. 
Then, labels in class-imbalanced CIFAR get flipped to the rest of the categories with the same probability. 
Clothing1M is a large-scale dataset with real-world noisy labels. 
In our experiment, clean training data is not used.

\paragraph{Experimental settings}
For CIFAR experiments, we use the PreAct ResNet-18~\cite{he2016identity} which is commonly used to benchmark label noise learning methods~\cite{li2020dividemix}. 
We train the network using SGD with a momentum of 0.9 for 300 epochs; warm-up 30 epochs for CIFAR-10 and 40 epochs for CIFAR-100. 
In the Clothing1M experiments, we use ResNet-50 with ImageNet pre-trained weights, following the previous work~\cite{li2020dividemix}. 
The warm-up period is 1 epoch for Clothing1M. 
$\tau$ is set as 0.6 for 90\% noise ratio and 0.5 for others. 
$\lambda_u$ is validated from $\{0, 25, 50, 150\}$. 
Generally, the hyperparameters setting for MixMatch is inherited from DivideMix without heavily tuning, because the SSL part is not our focus and can be replaced by other alternatives. 
We leverage MC-dropout~\cite{gal2016dropout} to estimate uncertainty, setting $T$ to 10 and the dropout rate to 0.3. 
The uncertainty ratio $r$ is set as 0.1 to obtain the final clean probability. 
We model aleatoric uncertainty for class-dependent and instance-dependent noise by 10 Monte Carlo samples on logits.

\paragraph{Baselines}
We compare our method (ULC) with multiple state-of-art approaches as follows: Co-teaching+~(CoT+)~\cite{yu2019does}, PENCIL~\cite{yi2019probabilistic}, Meta-Learning~(ML)~\cite{li2019learning}, M-correction~\cite{arazo2019unsupervised}, DivideMix~\cite{li2020dividemix}, and ELR+~\cite{liu2020early}.
All methods use the same architecture (PreAct ResNet-18 for CIFAR and ResNet-50 with ImageNet pretrained weights for Clothing1M).
A brief introduction to the baselines is in the appendix.

\subsection{Experimental Results}
We conduct experiments to verify the effectiveness of our approach and the baselines on the general class-balanced and class-imbalanced noisy data. 
The results of baselines on class-balanced CIFAR and Clothing1M are taken from \citet{li2020dividemix} and \citet{liu2020early}, and results on class-imbalanced CIFAR are obtained using publicly available code.
We put more experimental results, including the results for robustness to hyperparameters and the choice of SSL methods, to the appendix. 

\paragraph{Class-balanced synthetic CIFAR}
Table~\ref{tb:balance_result} shows a comparison among different methods on class-balanced CIFAR-10 and CIFAR-100.
The noise rate settings are the same with \citet{li2020dividemix}.
We report both the best~(indicating how well these methods learn from noisy samples) and the last~(indicating whether finally overfit) test accuracy following \citet{li2020dividemix}. 
ULC achieves comparable performance for low-level symmetric label noise, and outperforms state-of-the-art methods for high-level and asymmetric label noise. 

\begin{table}[t]
	\centering
	\resizebox{\columnwidth}{!}{
	\begin{tabular}{lc||*{4}c|c||*{4}c}
		\toprule  
		\multicolumn{2}{l||}{\multirow{2}{*}{Dataset}}&\multicolumn{5}{c||}{CIFAR-10}&\multicolumn{4}{c}{CIFAR-100}\\ 
		&&\multicolumn{4}{c|}{Sym.}&Asym.&\multicolumn{4}{c}{Sym.} \\
        \midrule
        \multicolumn{2}{l||}{Noise Rate}&20\%&50\%&80\%&90\%&40\%&20\%&50\%&80\%&90\% \\
        \midrule
        \multirow{2}{*}{CE}&best&86.8&79.4&62.9&42.7&85.0&62.0&46.7&19.9&19.9 \\
        &last&82.7&57.9&26.1&16.8&72.3&61.8&37.3&8.8&3.5 \\
        \midrule
        \multirow{2}{*}{CoT+}&best&89.5&85.7&67.4&47.9&-&65.6&51.8&27.9&13.7 \\
        &last&88.2&84.1&45.5&30.1&-&64.1&45.3&15.5&8.8 \\
        \midrule
        \multirow{2}{*}{PENCIL}&best&92.4&89.1&77.5&58.9&88.5&69.4&57.5&31.1&15.3 \\
        &last&92.0&88.7&76.5&58.2&88.1&68.1&56.4&20.7&8.8 \\
        \midrule
        \multirow{2}{*}{ML}&best&92.9&89.3&77.4&58.7&89.2&68.5&59.2&42.4&19.5 \\
        &last&92.0&88.8&76.1&58.3&88.6&67.7&58.0&40.1&14.3 \\
        \midrule
        \multirow{2}{*}{M-correction}&best&94.0&92.0&86.8&69.1&87.4&73.9&66.1&48.2&24.3 \\
        &last&93.8&91.9&86.6&68.7&86.3&73.4&65.4&47.6&20.5 \\
        \midrule
        \multirow{2}{*}{DivideMix}&best&\textbf{96.1}&94.6&93.2&76.0&93.4&77.3&74.6&60.2&31.5 \\
        &last&95.7&94.4&92.9&75.4&92.1&76.9&74.2&59.6&31.0 \\
        \midrule
        \multirow{2}{*}{ELR+}&best&95.8&94.8&93.3&78.7&93.0&\textbf{77.6}&73.6&60.8&33.4 \\
        &last&94.6&93.8&91.1&75.2&92.7&\textbf{77.5}&72.4&60.2&30.8 \\
        \midrule
        \multirow{2}{*}{ULC}&best&\textbf{96.1}&\textbf{95.2}&\textbf{94.0}&\textbf{86.4}&\textbf{94.6}&77.3&\textbf{74.9}&\textbf{61.2}&\textbf{34.5} \\
        &last&\textbf{95.9}&\textbf{94.7}&\textbf{93.2}&\textbf{85.8}&\textbf{94.1}&77.1&\textbf{74.3}&\textbf{60.8}&\textbf{34.1} \\
		\bottomrule  
	\end{tabular}
	}
	\vspace{\captionskip}
	\caption{Comparison with state-of-the-art methods in test accuracy on balanced CIFAR (\%).}
	\label{tb:balance_result}
	\vspace{\figureskip}
\end{table}

\paragraph{Class-imbalanced synthetic CIFAR}
Table~\ref{tb:imbalance_result} shows the results on class-imbalanced CIFAR-10 and CIFAR-100 with different levels of label noise. 
Note that the test sets here remain class-balanced, so the results can be compared with the class-balanced setting.
ULC consistently outperforms state-of-the-art methods across different imbalance ratios with different types of label noise.
Class-imbalance challenges the task of learning with noisy labels and all these previous methods work cannot perform well in this scenario. 
However, ULC shows robustness to this challenging scenario and works well even with only a few clean examples in minority classes (25 clean examples with 50\% symmetric noise under 1:10 class imbalance for CIFAR-100).

\begin{table}[t]
\renewcommand{\arraystretch}{\tabelarraystretch}
	\centering
	\resizebox{\columnwidth}{!}{
	\begin{tabular}{lc||cc|cc||cc|cc}
		\toprule  
		\multicolumn{2}{l||}{\multirow{2}{*}{Dataset}}&\multicolumn{4}{c||}{CIFAR-10}&\multicolumn{4}{c}{CIFAR-100}\\ 
		&&\multicolumn{2}{c|}{Sym. 20\%}&\multicolumn{2}{c||}{Sym. 50\%}&\multicolumn{2}{c|}{Sym.20\%}&\multicolumn{2}{c}{Sym.50\%} \\
        \midrule
        \multicolumn{2}{l||}{Resampling Ratio}&1:5&1:10&1:5&1:10&1:5&1:10&1:5&1:10 \\
        \midrule
        \multirow{2}{*}{CE}&best&87.1&85.4&72.0&68.4&64.9&61.4&48.2&43.0 \\
        &last&86.8&85.0&70.9&67.8&64.5&60.9&47.7&42.2 \\
        \midrule
        \multirow{2}{*}{CoT+}&best&82.5&76.7&71.9&53.2&49.2&39.9&30.6&29.2 \\
        &last&82.3&76.7&71.5&50.9&49.0&39.6&30.5&29.2 \\
        \midrule
        \multirow{2}{*}{PENCIL}&best&80.5&74.5&73.8&65.4&52.1&45.5&33.2&28.4 \\
        &last&80.0&73.0&73.5&65.2&51.1&43.3&31.9&25.2 \\
        \midrule
        \multirow{2}{*}{ML}&best&75.9&70.0&73.8&62.2&54.2&47.5&41.2&34.8 \\
        &last&74.6&68.0&61.6&54.0&51.1&44.4&35.2&30.3 \\
        \midrule
        \multirow{2}{*}{M-correction}&best&88.1&80.1&83.5&77.5&62.3&55.4&51.3&44.4 \\
        &last&87.0&77.3&83.1&76.8&62.2&54.7&50.0&44.1 \\
        \midrule
        \multirow{2}{*}{DivideMix}&best&93.9&74.8&85.5&66.9&65.0&51.2&56.9&44.2 \\
        &last&93.9&74.6&85.4&66.8&64.9&50.9&56.4&44.0 \\
        \midrule
        \multirow{2}{*}{ELR+}&best&88.2&79.8&82.7&65.5&59.6&49.9&53.4&44.7 \\
        &last&87.0&78.4&82.5&64.6&59.3&49.6&52.5&43.9 \\
        \midrule
        \multirow{2}{*}{ULC}&best&\textbf{95.0}&\textbf{93.8}&\textbf{94.9}&\textbf{92.5}&\textbf{75.5}&\textbf{72.8}&\textbf{71.6}&\textbf{57.2} \\
        &last&\textbf{94.9}&\textbf{92.6}&\textbf{94.7}&\textbf{92.2}&\textbf{75.1}&\textbf{72.3}&\textbf{70.9}&\textbf{56.7} \\
		\bottomrule  
	\end{tabular}
	}
	\vspace{\captionskip}
	\caption{Comparison with state-of-the-art methods in test accuracy on imbalanced CIFAR (\%).}
	\label{tb:imbalance_result}
	\vspace{\figureskip}
\end{table}

\paragraph{Clothing1M}
Table~\ref{tb:clothing1m} shows the results on the real-world noisy dataset Clothing1M. 
We report average accuracy as well as the standard deviation of five runs. 
ULC achieves comparable performance to state-of-the-art methods include DivideMix and ELR+.
The insignificant improvement may be due to two reasons.
Firstly, the estimated noise rate of Clothing1M is only 38.5\%~\cite{liu2020early}, secondly, the training set actually used for Clothing1M is nearly class-balanced~\cite{li2020dividemix}.

\begin{table}[t]
\renewcommand{\arraystretch}{\tabelarraystretch}
    \centering
	\resizebox{0.525\columnwidth}{!}{
    \begin{tabular}{l|c}
		\toprule  
        Method & Test Accuracy \\
        \midrule
        CE & 69.2 \\
        M-correction & 71.0 \\
        Meta-Learning & 73.5 \\
        PENCIL & 73.5 \\
        DivideMix & \textbf{74.8} \\
        ELR+ & \textbf{74.8} \\
        \midrule
        ULC & \textbf{74.9$\pm$0.2} \\
        \bottomrule
    \end{tabular}
    }
    \vspace{\captionskip}
	\caption{Comparison with state-of-the-art methods in test accuracy on Clothing1M (\%).}
	\label{tb:clothing1m}
    \vspace{\figureskip}
\end{table}

\subsection{Ablation Experiments}
Here, we study the individual components and their influence. 
Table~\ref{tb:ablation} summarizes the ablation study results. 
As we can see, each element provides an independent performance improvement. 

\paragraph{Class-specific label noise modeling}
Based on the above analysis, inter-class loss distribution discrepancy is the key to the degeneration of previous methods. 
As expected, the performance training without class-specific modeling rapidly breaks down with an increasing resampling ratio. 
Besides, we find that the influence of inter-class loss distribution discrepancy correlates to noise ratio. 
The performance deterioration training without class-specific modeling under 1:5 resampling with 20\% symmetric noise is modest, while it is very significant for 50\% symmetric noise.

\paragraph{Epistemic uncertainty-aware label noise modeling}
The performance consistently decreases without consideration of the effect of epistemic uncertainty on loss modeling, especially when the noise rate is high, or when the class imbalance property gets worse, as epistemic uncertainty-aware label noise modeling can provide a more reliable criterion for noisy sample discovery under series class imbalance or label noise.

\paragraph{Aleatoric uncertainty-aware learning}
When noise ratio and resampling ratio are both high-level, learning with aleatoric uncertainty is significantly beneficial to performance improvements. 
As noise modeling cannot completely remove label noise, and the network may still overfit residual noise.
This overfitting process is irreversible and may in turn affect the identification of label noise.
Learning with aleatoric uncertainty can alleviate final overfitting.

\begin{table}[t]
\renewcommand{\arraystretch}{\tabelarraystretch}
	\centering
	\resizebox{\columnwidth}{!}{
	\begin{tabular}{lc||cc|cc||cc|cc}
		\toprule  
		\multicolumn{2}{l||}{\multirow{2}{*}{Dataset}}&\multicolumn{4}{c||}{CIFAR-10}&\multicolumn{4}{c}{CIFAR-100}\\ 
		&&\multicolumn{2}{c|}{Sym. 20\%}&\multicolumn{2}{c||}{Sym. 50\%}&\multicolumn{2}{c|}{Sym.20\%}&\multicolumn{2}{c}{Sym.50\%} \\
        \midrule
        \multicolumn{2}{l||}{Resampling Ratio}&1:5&1:10&1:5&1:10&1:5&1:10&1:5&1:10 \\
        \midrule
        \multirow{2}{*}{ULC}&best&\textbf{95.0}&\textbf{93.8}&\textbf{94.9}&\textbf{92.5}&\textbf{75.5}&\textbf{72.8}&\textbf{71.6}&\textbf{57.2} \\
        &last&\textbf{94.9}&\textbf{92.6}&\textbf{94.7}&\textbf{92.2}&\textbf{75.1}&\textbf{72.3}&\textbf{70.9}&\textbf{56.7} \\
        \midrule
        \multirow{2}{*}{ULC w/o CSM}&best&94.2&82.5&88.5&70.5&67.1&59.4&61.3&50.2 \\
        &last&93.8&81.3&87.9&69.8&66.2&58.7&60.5&49.8 \\
        \midrule
        \multirow{2}{*}{ULC w/o EUM}&best&94.1&90.7&91.3&89.9&68.0&67.2&65.9&53.2 \\
        &last&93.8&90.1&90.8&88.7&67.5&66.9&65.4&52.6 \\
        \midrule
        \multirow{2}{*}{ULC w/o AUL}&best&94.2&91.8&92.5&91.2&70.2&66.5&68.3&52.2 \\
        &last&94.0&91.5&91.7&90.5&69.7&66.1&67.2&51.6 \\
		\bottomrule  
	\end{tabular}
	}
	\vspace{\captionskip}
	\caption{Ablation study results on imbalanced CIFAR (\%). CSM refers to class-specific noise modeling, EUM refers to epistemic uncertainty-aware noise modeling, and AUL refers to aleatoric uncertainty-aware learning.}
	\label{tb:ablation}
	\vspace{\figureskip}
\end{table}

\section{Conclusion}
In this work, we propose an uncertainty-aware label correction framework~(ULC) to adapt for learning with noisy labels on class-imbalanced data. 
We mainly focused on overcoming inter-class loss distribution discrepancy and
alleviating the effect of epistemic uncertainty and aleatoric uncertainty. 
We model label noise incorporating class-specific loss distribution with estimated epistemic uncertainty to identify more confident clean samples from noisy samples. 
In the process of learning with corrected labels, we formulate aleatoric uncertainty induced by residual label noise as logit corruption into loss to alleviate overfitting in majority classes and during the late learning phase. 
Experiments show that ULC yields strong results on different benchmarks despite the increasing noise ratio and resampling ratio. 
In the future, we will investigate the facilitation effect of uncertainty on learning with open-set label noise, considering that out-of-domain data can be viewed as a reason leading to epistemic uncertainty.

\clearpage
\bibliography{refer}

\clearpage
\appendix

\section{Detailed Information of the Proposed Method}
We describe more detailed information in this appendix.

\subsection{Uncertainty in Deep Learning}
The possibilities in model parameters lead to epistemic uncertainty, while noisy data leads to aleatoric uncertainty. 
Epistemic uncertainty and aleatoric uncertainty induce misleading prediction~\cite{northcutt2021confident}.
BNNs~\cite{gal2016dropout} make it possible for uncertainty estimation in DNNs. 
Epistemic uncertainty can be modeled by placing a prior distribution over the model's parameters, $p(W)$. 
With the formulation, BNNs aims to infer the posterior $p(W|\mathcal{X, \widetilde Y})$ and predict the marginal probability $p(\widetilde{\mathcal{Y}}|\mathcal{X})$. 
Based on variational inference methods, different approximations with a surrogate distribution $q_\theta(W)$, parameterized by $\theta$, have been developed to fit the posterior. 
Considering $q_\theta(W)$ to be the Dropout distribution~\cite{srivastava2014dropout}, $T$ sampled masked model weights $\{W_t\}_{t=1}^T \sim q_\theta(W)$ can be obtained by performing $T$ stochastic forward passes through the network with dropout enabled. 
Accordingly, the prediction can be approximated using Monte Carlo integration as follows:
\begin{equation}
\small
  p(y|x) = \int p(y|x,W)q_\theta(W)dW \approx \frac{1}{T}\sum_{t=1}^{T}\textrm{softmax}(f(x;W_t))\,.
\end{equation}

The epistemic uncertainty can be summarised using the entropy of the probability:
\begin{equation}
\label{eq:uncertainty}
\small
    \epsilon(\hat y_i) = -\sum_{j=1}^{C}(\frac{1}{T}\sum_{t=1}^{T}{\hat y}_{ijt}) \log  (\frac{1}{T}\sum_{t=1}^{T}{\hat y}_{ijt})\,.
\end{equation}

On the other side, the aleatoric uncertainty can be modeled by placing a prior distribution over the output logit of the model, for example, a Gaussian distribution: $v\sim\mathcal{N}(f(x),\sigma)$.
Readers may refer to \citet{kendall2017uncertainties} for more details.

\subsection{Summarization of Notations}
We summarize the notations used in the paper in Table~\ref{tb:notations}.

\begin{table}[th]
    \centering
	\resizebox{\columnwidth}{!}{
    \begin{tabular}{r|p{9.35cm}}
		\toprule  
        Notation & Definition \\
        \midrule
        $\mathcal{X}$ & The set of training examples \\
        $x_i$ & The $i^{th}$ training example \\
        $\widetilde{\mathcal{Y}}$ & The set of observed noisy labels \\
        $\widetilde{y_i}$ & The observed noisy label corresponding to the $i^{th}$ training example, $y_i \in \{0, 1\}$ \\
        $y_i^*$ & The latent true label corresponding to $\widetilde{y}_i$ \\
        $\widetilde{y}_{ij}$ & The $j^{th}$ element of $\widetilde{y}_{i}$ \\
        ${y}_{ij}^*$ & The $j^{th}$ element of ${y}_{i}^*$ \\
        $f$ & DNN model parameterized with $W$ \\
        $W$ & The set of trainable model parameters\\
        $v_i$ & The output logits for $x_i$ \\
        $\widehat{y}_i$ & The predicted probability of the $i^{th} example$ \\
        $\ell$ & The cross-entropy loss on the training data \\
        $\ell_i$ & The cross-entropy loss for the $i^{th}$ sample \\
        $y_i$ & The corrected label corresponding to $\widetilde{y}_i$ \\
        $\mathcal{Y}$ & The set of corrected labels \\
        $\widehat{\mathcal{X}}$ & The set of examples identified as clean \\
        $\widehat{\mathcal{U}}$ & The set of examples identified as noise \\
        $\ell_c$ & The corrected loss \\
        $\ell_x$ & The supervised loss for examples identified as clean \\
        $\ell_u$ & The unsupervised loss for examples identified as noise \\
        $\lambda_u$ & The hyperparameter for weighting the unsupervised loss \\
        $\ell_\textrm{reg}$ & The regularization term in the corrected loss \\
        $\lambda_\textrm{reg}$ & The hyperparameter for weighting the regularization term \\
        $y_i^j$ & The corrected label for the $i^{th}$ example in the $j^{th}$ iteration \\
        $\mathcal{L}$ & The set of cross-entropy loss values \\
        $\mathcal{L}_i$ & Subset of cross-entropy loss values for examples with observed noisy label $i$ \\
        $\mathcal{L}^n$ & The set of cross-entropy loss values in the $n^{th}$ update \\
        $\mathcal{Y}^n$ & The set of corrected labels in the $n^{th}$ update \\
        $h$ & Function used to rectify labels based on model predictions. \\
        $\epsilon(\widehat{y}_i)$ & Estimated epistemic uncertainty for $\widehat{y}_i$ \\
        $\widehat{y}_{ijt}$ & The $j^{th}$ element of the $t^{th}$ stochastic forward prediction for the $i^{th}$ example \\
        $\widehat{y}_{it}$ & The $t^{th}$ sampled prediction for the $i^{th}$ example \\
        $q_\theta(W)$ & Surrogate distribution with parameters $\theta$ for the posterior probability of $W$ \\
        ${\{W_t\}}_{t=1}^T$ & $T$ sampled model weights from $q_\theta(W)$ \\
        $\omega_i$ & Estimate clean probability of $\widetilde{y}_i$, namely the probability of $\widetilde{y}_i = y_i^*$ \\
        $\mu_{j0},\mu_{j1}$ & The mean of GMM modeled on loss values, $\mu_{j0} \le \mu_{j1}$ \\
        $r$ & The tuneable hyperparameter to adjust the impact of uncertainty on noise modeling \\
        $\delta_{\mathcal{Y}}^{x_i}$ & Instance-dependent and class-dependent random noise injected into logits \\
        $\delta^{x_i}$ & Instance-dependent Gaussian noise \\
        $\delta$ & Class-dependent Gaussian noise \\
        $\sigma^{x_i}$ & Variance of instance-dependent noise $\delta^{x_i}$ \\
        $\sigma$ & Variance of class-dependent noise $\delta$ \\
        $\mathcal{X}'$ & Mixed-up augmented labeled training data \\
        $\mathcal{U}'$ & Mixed-up augmented unlabeled training data \\
        \bottomrule
    \end{tabular}
    }
    \vspace{\captionskip}
	\caption{Notation used in ULC}
	\label{tb:notations}
    \vspace{\figureskip}
\end{table}

\subsection{Algorithm Pseudocode for ULC}
Algorithm~\ref{alg:ulc} presents the step-by-step procedure of the proposed ULC.

\begin{algorithm}[t!]
\caption{Uncertainty-aware Label Correction}
\label{alg:ulc}
\textbf{Input}: Noisy training data $(\mathcal{X}, \mathcal{\widetilde Y})$\\
\textbf{Hyper-parameter}: Uncertainty ratio $r$, clean probability threshold $\tau$, MC sampling times $T$, hyper-parameters $M,\lambda_u,\alpha$ for MixMatch\\
\textbf{Output}: Model parameters $W^{(1),(2)}$, $\sigma^{\mathcal X(1),(2)}$, $\sigma^{(1),(2)}$, $\mathcal Y^{(1),(2)}$
\begin{algorithmic}[1] 
\small
\STATE $(W,\sigma^{\mathcal X},\sigma)^{(1),(2)} \gets \textrm{WarmUp}(\mathcal{X}, \mathcal{\widetilde Y})$.
\WHILE{$e \leq \textrm{MaxEpoch}$}
\STATE Infer integrated predictions $\mathcal{\widehat Y}^{(1),(2)}$, loss values $\mathcal L^{(1),(2)}$, epistemic uncertainty $\epsilon^{(1),(2)}(\widehat{\mathcal{Y}})$.
\FOR{$c=0$ to $C$}
\STATE /* Class-specific noise modeling */
\STATE $p^{(2)}(\mu_{j0}|\ell_i) \gets \textrm{GMM}(\mathcal{L}^{(1)}_c)$\\
\STATE $p^{(1)}(\mu_{j0}|\ell_i) \gets \textrm{GMM}(\mathcal{L}^{(2)}_c)$
\FOR{$k=1,2$}
\STATE /* Modulating noise modeling on loss values by epistemic uncertainty */
\STATE $\omega^{(k)}_i \gets p^{(k)}(\mu_{j0}|\ell_i)^{1-r}(1-\epsilon^{(k)}(\widehat{y}_i))^{r}$
\STATE $y^{(k)}_i \gets \omega^{(k)}_i \widetilde{y}_i + (1-\omega^{(k)}_i)\widehat{y}_i$
\ENDFOR
\ENDFOR
\FOR{$k=1,2$}
\STATE $\mathcal{\widehat X}^{(k)}_e \gets \left\{ (x_i,y^{(k)}_i)|\omega_i \ge \tau, \forall(x_i,y^{(k)}_i,\omega_i) \in (\mathcal X, \mathcal Y, \omega^{(k)}) \right\}$.
\STATE $\mathcal{\widehat U}^{(k)}_e \gets \left\{x_i| \omega_i < \tau, \forall(x_i,\omega_i)\in (\mathcal X,\omega^{(k)}) \right\}$.
\FOR{$iter=1$ to $num_iter$}
\STATE From $\mathcal{\widehat X}^{(k)}_e$, draw a mini-batch $\mathcal{\widehat X}_B$
\STATE From $\mathcal{\widehat U}^{(k)}_e$, draw a mini-batch $\mathcal{\widehat U}_B$
\STATE $\mathcal X'_B,\mathcal U'_B \gets \textrm{MixMatch}(\mathcal{\widehat X}_B,\mathcal{\widehat U}_B)$.
\STATE $\mathcal X'_B=\{(x_b, y_{x_b});b\in (1,\dots,B) \}$
\STATE $\mathcal U'_B=\{(u_b, y_{u_b});b\in (1,\dots,B) \}$
\FOR{$b=1$ to $B$}
\FOR{$t=1$ to $T$}
\STATE /* Sampling over logit vector */
\STATE $\widehat v_{{x_b}t} \gets (I + \delta_t)f_W(x_b) + \delta_t^{x_b}$.
\STATE $\widehat v_{{u_b}t} \gets (I + \delta_t)f_W(u_b) + \delta_t^{u_b}$.
\ENDFOR
\STATE $\widehat y_{x_b} \gets \frac{1}{T}\sum_{t}\textrm{softmax}(\widehat v_{{x_b}t})$.
\STATE $\widehat y_{u_b} \gets \frac{1}{T}\sum_{t}\textrm{softmax}(\widehat v_{{u_b}t})$.
\STATE $\ell_{c_b} \gets \ell_x(\widehat y_{x_b}, y_{x_b}) + \lambda_u \ell_u(\widehat y_{u_b}, y_{u_b})$.
\ENDFOR
\STATE $W^{(k)}=SGD(\frac{1}{B}\sum_{b=1}^{B}\ell_{c_b},W^{(k)})$.
\ENDFOR
\ENDFOR
\ENDWHILE
\end{algorithmic}
\end{algorithm}


\section{Additional Details and Experimental Results}
We first give a brief introduction to the baselines, and then discuss the robustness of ULC to hyperparameters, and the influence of the different choices of SSL methods. 
We implement the proposed method by PyTorch 1.2.0 with default initialization and conduct all the experiments on a single Nvidia P40 GPU.

\subsection{Brief Introduction to the Baselines}
Here, we introduce some of the most recent state-of-the-art methods for comparison.
Co-teaching+~\cite{yu2019does} maintains two networks simultaneously to teach each other and improves Co-teaching by exploiting prediction disagreement to keep the two networks diverged.  PENCIL~\cite{yi2019probabilistic} adopts label probability distributions to supervise network learning and updates both network parameters and label estimations as label distributions. 
Meta-Learning~\cite{li2019learning} performs a meta-learning update before conventional training on synthetic noisy labels to prevent overfitting to artificially generated label noise. 
M-correction~\cite{arazo2019unsupervised} models per-sample loss with a two-component BMM to carry out sample selection, correct labels via a convex combination of the noisy labels and the soft or hard labels, and applies MixUp augmentation to enhance performance. 
DivideMix~\cite{li2020dividemix} dynamically fits a two-component GMM on per-sample loss distribution to distinguish clean samples from noisy ones, and utilizes MixMatch to addresses learning with label noise in a semi-supervised manner. 
ELR+~\cite{liu2020early} leverages SSL to produce target probabilities based on the outputs probabilities, exploits the memorization effect on early learning via regularization, and use a combination of temporal ensembling, weight averaging, and MixUp augmentation to improve performance. 
Both DivideMix and ELR+ train two networks to avoid confirmation bias.

\subsection{Robustness to Hyperparameters}
\begin{figure}[t]
    \centering
    \includegraphics[width=0.465\columnwidth]{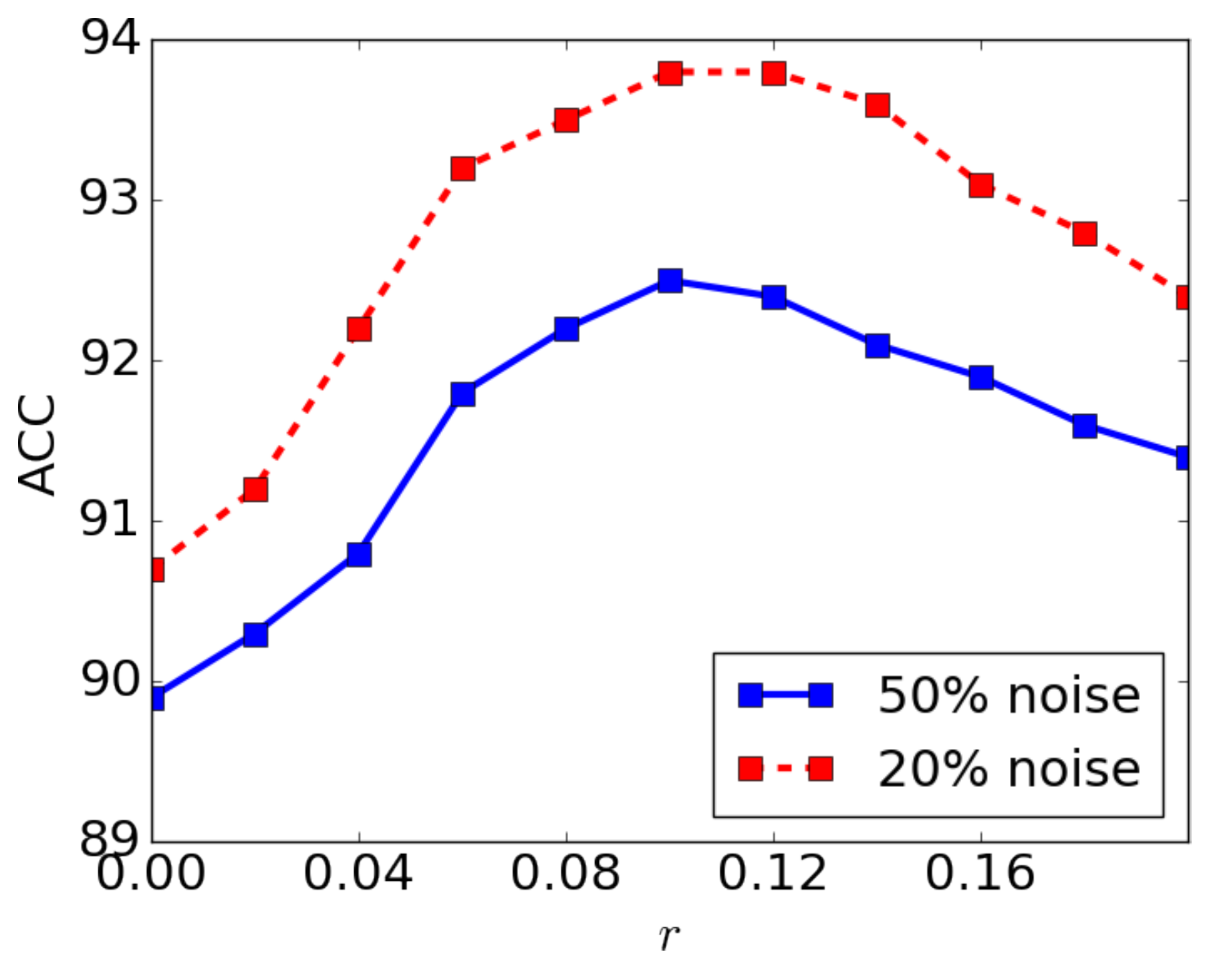}
    \vspace{\captionskip}
    \caption{Testing accuracy (\%) with $r$ ranging from 0.0 to 0.2 on 1:10 class-imbalanced CIFAR-10 with 20\% and 50\% symmetric noise.}
    \label{Fig4}
    \vspace{\figureskip}
\end{figure}

ULC introduces a new hyperparameter $r$ to adjust the impact of uncertainty on label noise modeling. 
We select $r$ based on the CIFAR-10 test set, training under 1:10 class imbalance with 50\% symmetric label noise. 
We set $r=0.1$, and find it works well across datasets. 
Moreover, we find ULC is relatively robust to this hyperparameter, and using $0.05 \le r \le 0.15$ leads to comparable performance. 
Fig~\ref{Fig4} shows the test accuracy produced as $r$ varies on 1:10 class-imbalanced CIFAR-10 with 20\% and 50\% symmetric noise.

\subsection{The Choice of SSL Methods}
To reuse samples that are identified as noise in the label noise modeling process, we leverage SSL methods to attach an unsupervised loss on these unlabeled samples. 
For a fair comparison to DivideMix, we use MixMatch while other methods are also applicable. 
We find that Fixmatch~\cite{sohn2020fixmatch} as an alternative works better, and achieves test accuracy 75.2$\pm$0.2\%~(mean $\pm$ standard deviation over five runs) on Clothing1M. 
However, the exploration of SSL methods and the corresponding augmentation strategies are not the focus of this paper. 

\subsection{Training Time Analysis}
Our proposed ULC has theoretically similar training time to DivideMix. 
The process of uncertainty estimation and logit corruption introduces marginal computation cost. 
We only estimate epistemic uncertainty in the label noise modeling procedure and perform $T$ stochastic forward passes through the top layers with dropout. 
We sample on $C\times C$ matrices and $C\times 1$ vectors to model aleatoric uncertainty and perform logit corruption. 
The additional computation cost is modest compared to network inference.

\end{document}